\documentclass{article}
\usepackage{spconf,amsmath,graphicx}
\usepackage[utf8]{inputenc}
\usepackage{wrapfig}
\usepackage{subcaption}
\usepackage{longtable,booktabs}
\usepackage{tabularx}
\usepackage{makecell}
\usepackage{environ}
\usepackage{tikz}
\usepackage{float}
\usepackage{enumitem}
\usepackage[font=small]{caption}
\usepackage[T1]{fontenc}  
\usepackage{textcomp}     
\usepackage{lipsum}

\let\OLDthebibliography\thebibliography
\renewcommand\thebibliography[1]{
  \OLDthebibliography{#1}
  \setlength{\parskip}{0.5ex}
  \setlength{\itemsep}{0pt plus 0.3ex}
}

\NewEnviron{elaboration}{
\begin{tikzpicture}
\node[rectangle,minimum width=0.4\textwidth] (m) {\begin{minipage}{0.46\textwidth}\BODY\end{minipage}};
\draw[dashed] (m.south west) rectangle (m.north east);
\end{tikzpicture}
}

\title{Pre-Training for Query Rewriting \\ in a Spoken Language Understanding System}
\twoauthors
 {Zheng Chen \sthanks{This work was completed while the first author was an intern at Amazon.}}
  {Drexel University, \\ College of Computing \& Informatics}
 {\parbox{150pt}{\centering Xing Fan, Yuan Ling \newline Lambert Mathias, Chenlei Guo}}
 {Amazon.com, Inc. USA}
\begin{document}
%
\maketitle
\begin{abstract}
Query rewriting (QR) is an increasingly important technique to reduce customer friction caused by errors in a spoken language understanding pipeline, where the errors originate from various sources such as speech recognition errors, language understanding errors or entity resolution errors. In this work, we first propose a neural-retrieval based approach for query rewriting. Then, inspired by the wide success of pre-trained contextual language embeddings, and also as a way to compensate for insufficient QR training data, we propose a language-modeling (LM) based approach to pre-train query embeddings on historical user conversation data with a voice assistant. In addition, we propose to use the NLU hypotheses generated by the language understanding system to augment the pre-training. Our experiments show pre-training provides rich prior information and help the QR task achieve strong performance. We also show joint pre-training with NLU hypotheses has further benefit. Finally, after pre-training, we find a small set of rewrite pairs is enough to fine-tune the QR model to outperform a strong baseline by full training on all QR training data.
\end{abstract}
\vspace{-1mm}
\section{Introduction}
\vspace{-1mm}
\label{sec_intro}
Spoken language understanding (SLU) systems are responsible for extracting semantic information from an input voice query. Advances in SLU have been central to user adoption of voice assistants like Alexa, Google Home and Siri to accomplish their daily tasks. Many SLU systems are modeled as two separate components in a pipeline - an automatic speech recognition (ASR) system responsible for converting audio input to texts, and a natural language understanding (NLU) component that extracts meaning\footnote{They are typically represented as domains, intents and slots.} for downstream applications to act on. Each of these components in the pipeline can introduce errors - ASR can misrecognize text due to background noise, speaker accents and the NLU component can misinterpret the semantics due to ambiguity as well as errors from ASR. These errors accumulate and introduce {\it friction} in the conversation, resulting in users rephrasing their requests or abandoning their tasks. Hence, reducing friction, we posit, is central to the adoption of spoken language interfaces.

There is a large body of work focusing on friction reduction on the ASR component or the NLU component, e.g. ~\cite{shokouhi2016did, shivakumar2019learning, ponnusamy2019feedback, serdyuk2018towards}. A more scalable approach, which does not require modifying the ASR or NLU components, is to introduce a \textit{query rewriting} (QR) module that takes in the ASR transcript and rewrites it before sending it to the downstream NLU sub-system. This has been successful in web search~\cite{shokouhi2014mobile,dehghani2017learning, riezler2010query, he2016learning, guo2019spelling}, aided by the success of deep learning seq2seq~\cite{sutskever2014sequence} architectures that handle longer context and require little feature engineering. In this paper, we propose a neural retrieval based QR approach to reduce user friction caused by the various aforementioned reasons. We define the ASR output of a given utterance spoken by a user as the \textit{query}, and we aim at replacing the original query by a reformulated query, i.e. the \textit{rewrite}, that could improve the overall success rate. Taking Figure \ref{fig_sess_example} for example, given the initial defective query "play lonely eyes old time road", our goal of QR is to replace it by a \textit{good rewrite} "play lil nas x. old town road" with the correct artist name and song name so that the voice assistant can act on this to accomplish the user task. 

{
\setlength\intextsep{-1pt}
\begin{figure}
  \centering
  \includegraphics[width=0.48\textwidth]{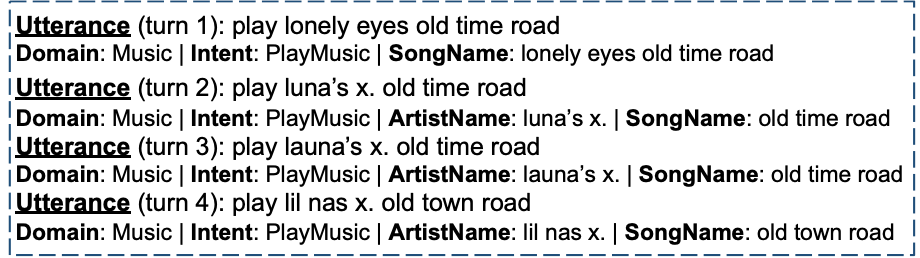}
  \vspace{-6mm}
  \caption{A session of utterances and their NLU hypotheses.}
  \label{fig_sess_example}
  \vspace{-2.0em}
\end{figure}
}

A neural QR model can be trained using labeled training data in the form of pairs of the original query and the rewrite. Such data set is expensive to obtain, yet a neural QR model typically needs a large number of rewrite pairs for training in order to achieve generalized performance (i.e. performance  not biased toward particular types of rewrites). Recent work on training contextual word representations~\cite{devlin2018bert, peters2018deep,radford2018improving} on large diverse text corpora and then fine-tuning on task specific labeled data~\cite{Howard2018UniversalLM} has helped alleviate the large data requirements. Pre-training works best when the data match the task. For example, ~\cite{henderson2019training} showed that pre-training on actual conversations and related corpora can benefit the task of dialogue response selection. Inspired by these work, we also propose to pre-train query embeddings, taking advantage of existing SLU system's large amount of historical dialogue sessions and users' \textit{implicit feedback}. We define one \textit{session} as a sequence of utterances spoken by one user during an engagement with the voice assistant where the time delay between any two consecutive utterances is at most 45 seconds. In addition, each query in a session is associated with its \textit{NLU hypothesis} generated by the NLU component, labeling the domain, intent and slots (slot types/values) for the query. 

For example, Figure \ref{fig_sess_example} shows a session where a user gave implicit feedback by repeating the utterance until the ASR finally output the correct query for the intended song. Below each query is their NLU hypotheses. The \textit{domain} is the general topic of an utterance, i.e. "Music". The \textit{intent} reflects the action the user wants to take, i.e. "PlayMusic"; the \textit{slot-types/values} are results of entity labeling, i.e. "SongName" is the  entity type with "old town road" as the slot value. 

In this paper, we first propose a neural retrieval based QR approach which includes design of a query embedder built on top of a pre-trained contextual word/subword embedder. Then we propose to make use of the historical query data and pre-train the query embedder. We also show how we can leverage additional benefits from the NLU hypotheses generated by the NLU system as the regularization or weak supervision. Experimental results and discussion will be included afterwards to show the effectiveness of our approach.

\vspace{-1mm}
\section{Methodology}
\label{sec:pagestyle}
\vspace{-1mm}
In this section, we first describe our neural retrieval based QR baseline system, consiting of a query embedder and an index of pre-defined rewrite candidates. Then we will present our method to pre-train the query embedder with the large amount of readily available historical sessions and how to incorporate NLU hypotheses in the pre-training. 
\vspace{-2mm}
\subsection{Query Rewrite}
\label{sec_baseline}
\vspace{-1mm}
We frame the QR problem as an information retrieval (IR) task. Given a query and millions of indexed rewrite candidates, we select the most relevant candidate as the query's rewrite. Many state-of-the-art approaches in paraphrase identification \cite{he2015multi, lan2018neural} are less practical for the QR task given a large number of candidates. Also for this paper we do not include the seq2seq approach like \cite{dehghani2017learning}; in our practice it is found harder to ensure good quality of generated rewrites than historical rewrites, and so left as future work. Instead, we design a neural retrieval system based on query embedding, where a neural encoder learns to capture latent syntactic and semantic information from the given query specifically for the QR task. This way, we can compute the fixed size vector representation for the large number of rewrite candidates offline. The neural encoder can be trained that the query embedding is close to its corresponding good rewrites in the projected space. 
\begin{figure}
  \begin{center}
    \includegraphics[width=0.45\textwidth]{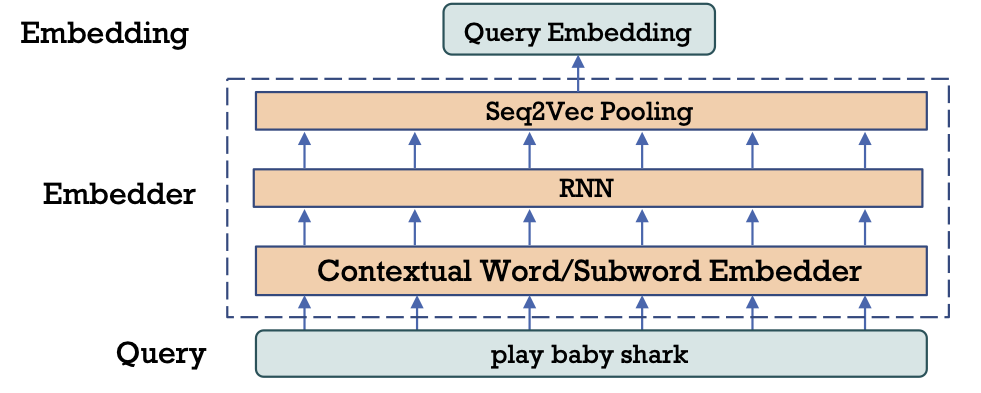}
  \end{center}
  \vspace{-4mm}
  \caption{The proposed query embedder architecture.}
  \vspace{-1.5em}
  \label{fig2}
\end{figure}

We propose an encoder architecture in Figure \ref{fig2} referred as \textit{embedder}. Given an input query, we first extract the token level representation by feeding the query into a pre-trained contextual word/subword embedding models, e.g. ELMo, BERT. This step helps to leverage the rich prior semantic information learned in these models. Then we add one RNN layer such like the bidirectional-LSTM, before we merge the token level representation into an query level embedding though a sequence-to-vector (seq2vec) pooling layer. The seq2vec layer can be implemented by mean-pooling, max-pooling, or self-attention, etc. We formally define this series of nonlinear transformation as
\small
\vspace{-0.7em}
\begin{equation}
\mathbf{u} = \text{Emb}(u)
\vspace{-0.7em}
\label{eq_embedding}
\end{equation}
\noindent
\normalsize
\noindent
where $u$ is a query and $\mathbf{u}$ is its final query embedding. In order to measure the similarity between $\mathbf{u}$ and $\mathbf{u}'$, the embedding of two queries $u, u'$ in the projected space, we use the \textit{scaled cosine distance} as in Eq.(\ref{eq_distance}), where $\alpha$ is a scaling constant and we empirically choose $\alpha=16$ and $\textit{linear}_1, \textit{linear}_2$ are linear transformations on top of $\mathbf{u}$. In comparison to the commonly used dot product \cite{pennington2014glove, mikolov2013distributed}, Eq.(\ref{eq_distance}) can be viewed as kind of normalized dot product, and it only encourages smaller angles but not longer embeddings that could be geometrically far apart from each other in the Euclidean space.

\small
\vspace{-0.7em}
\begin{equation}
\Delta(\mathbf{u}, \mathbf{u}') = \alpha \times \operatorname{cos}\left(\textit{linear}_1(\mathbf{u}),\textit{linear}_2(\mathbf{u}')\right)
\label{eq_distance}
\end{equation}
\noindent
\normalsize

\noindent
For training, given a pair $(u, u')$, where $u$ is the original query and $u'$ is its rewrite, we calculate the probability of $p(u'|u)$ as
\vspace{-0.3em}
\small
\begin{equation}
p\left( u'|u \right) = \frac{\exp\left\{ \Delta\left( \text{Emb}(u), \text{Emb}(u') \right) \right\}}{\sum_{\tilde{u} \in R}^{}{\exp\left\{ \Delta\left( \text{Emb}(u),\text{Emb}({\tilde{u}}) \right) \right\}}},
\label{eq_pred_layer2}
\end{equation}
\noindent
\normalsize
where ideally $R$ should be all rewrite candidates. In this paper, for training efficient, we construct it from each mini-batch, i.e. letting $R$ be the set of rewrite candidates in a mini-batch. We then use the cross entropy as the loss function.

Besides the neural encoder, the other key component of the retrieval system is the k-Nearest Neighbor (kNN) index. The indexed rewrite candidates come from a set of pre-defined high-precision rewrite pairs selected from Alexa's historical data with high impression \cite{ponnusamy2019feedback}. We index both the original queries and the rewrites from these pairs, which results an index of millions of rewrite candidates. All rewrite candidates are encoded offline using the neural encoder and added to the kNN index implemented using FAISS \cite{johnson2019billion}. At retrieval time, the query is encoded and passed to FAISS to retrieve the top-k relevant rewrites from the projected space. 

\vspace{-2mm}
\subsection{Pre-training of Query Embedding}
\label{sec_approach_pre_train}
\normalsize
\vspace{-1mm}
The neural QR model requires a considerable amount of training data to achieve good and generalized performance. However, annotated QR data is typically not available. One intuitive way to alleviate this problem is to automatically construct a set of likely rewrite pairs - using a rephrase detection model to process the historical sessions and select pairs assigned high scores by the detection model. However, the number of selected pairs by rephrase detection will only be a small portion of all historical data. The majority of data is wasted and we might risk bias in the selected pairs. This method will serve as the strong baseline in our experiments.

Inspired by the success of pre-trained contextual word or subword embeddings, we propose to add a pre-training process before we fine-tune the model with query rewrite pairs. In particular, we extend the pre-training process at query level by leveraging a large quantity of query history within a session. In this section, we describe the query embedding first and then extend it to incorporate the NLU hypothesis embedding as a multi-task training serves for regularization.

\textbf{Pre-Training using Language Modeling}. The user query history of an existing SLU system contains rich and helpful user implicit feedback. Pre-trained model using these data should learn to incorporate these rich prior information for the purpose of generalization. In this paper, we treat the query embedding pre-training problem as a language modeling (LM) problem at the query level, where given the current query $u_{t}^{(s)}$ of a session $s$, we try to evaluate $p(u_{t+1}^{(s)}|u_t^{(s)})$, i.e. predicting the query at the next turn in the same session. We adapt the quick thoughts (QT) framework from \cite{logeswaran2018efficient} to train this language model. For the encoder part, we take the same architecture as described in Section \ref{sec_baseline} and we follow the same training objective function as in \cite{logeswaran2018efficient}.

\textbf{Joint Training with NLU Hypotheses}. The NLU component in a SLU system provides a semi-structured semantic representation for queries, where queries of various text forms but the same semantics can be grouped together through the same NLU hypothesis. For example, "could you please play imagine dragons", "turn on imagine dragons", "play songs from imagine dragons" carry the same semantics and have the same NLU hypothesis, but their texts are different. Intuitively, augmenting the query texts with the less noisy NLU hypotheses could be helpful. Another reason to involve NLU hypotheses in the pre-training is that they are not available during runtime, because our QR module sits between ASR and NLU in the SLU system. Thus, the joint pre-training will preserve the information from the NLU hypothesis into the query embeddings and make use of them during runtime. 

In this session, we propose a joint trained language models by both query texts and their NLU hypotheses. We aim to project the query and its NLU hypothesis to the same space, therefore serves as a way of regularization or weak supervision. Let $u_t^{(s)}, h_t^{(s)}$ be the current utterance and hypothesis in session $s$. Instead of just having a task of $p(u_{t+1}^{(s)}|u_t^{(s)})$, we use four prediction tasks for joint pre-training: $p(u_{t+1}^{(s)}|u_t^{(s)})$, $p(h_{t+1}^{(s)}|u_t^{(s)})$, $p(u_{t+1}^{(s)}|h_t^{(s)})$ and $p(h_{t+1}^{(s)}|h_t^{(s)})$. 




For this joint training, we continue to use the model architecture shown in Fig. \ref{fig2}, except now the input can be either the query text form or the query NLU hypothesis form. We process the query text in the same way as before. For the NLU hypothesis, we concatenate the domain, intent, slot type and slot value and serialize it into a word sequence, which will then be processed in the same way as the query text form and feed to the embedder. For the training objective function, the four tasks are added together without task-specific weights.

\vspace{-3mm}
\section{Experiments}
\label{sec_experiments}
\vspace{-3mm}
\subsection{Data}
\vspace{-1mm}
\textbf{Training Sets}. We construct two separate data sets for model training. The one used to pre-train the utterance/hypothesis embedding is referred to as the “pre-training set”, and the other one used to train the baseline or fine-tune the pre-trained model is called the "QR set".

For the pre-training set, we extract all historical sessions where each session contains at least two consecutive queries across all available users in a given time range, i.e. a few weeks. We further remove sessions ended up with an \textit{intent} indicating unsuccessful customer experience, i.e. "CancelIntent" or "StopIntent". Through this process, we collect a pre-training set of 11 million sessions with about 30 million utterances, where each session contains various number of customer queries, mostly ranging from 2 to 6.

Different from the pre-training set, the fine-tuning set is generated using an existing rephrase detection model pipeline\footnote{The rephrase detection model is a BERT based model fine tuned with annotated rephrase pairs \cite{aguilar2019knowledge} to generate a probability that the second query is a rephrase form of the first query.}. Within a session, we select query pair that results a high confidence score from the rephrase detection model. The selected query pair is with the form of $(u,u')$, where with high confidence that the second query $u'$ is a corrected form of the first unsuccessful query $u$. This process gives a fine-tune set of 2.2 million utterance pairs. The following are examples of some query rewrite pairs in the fine-tune set:
{\small
\begin{itemize}[topsep=0pt,itemsep=-1ex,partopsep=1ex,parsep=1ex]
  \item \textit{play ambient mean} $\rightarrow$ \textit{play envy me}, \\ where the rewrite corrects the ASR error;
  \item \textit{play blues radio news} $\to$ \textit{play blue news radio}, \\ where the rewrite rearranges the words for better clarity;
  \item \textit{play relax music} $\to$ \textit{play relaxing music from my playlist}, \\ where the rewrite specifies query details;
\end{itemize}
}
\noindent
\textbf{Test Set}. In order to evaluate our QR approach, we first follow a procedure similar to how we construct the QR set, but the test data are randomly sampled from a time period immediately following the period from which the training data was drawn. Then we have manually identified the true rephrase pairs out of the sampled $(u, u')$. The one confirmed as true rephrase pair will be included in the QR evaluation set. This way, we derived an evaluation set with 16K $(u, u')$ pairs. 

\vspace{-2mm}
\subsection{Experimental Setup}
\label{sec_exp_setup}
\vspace{-1mm}
To evaluate the QR performance, we compare the retrieved rewrite candidate's NLU hypothesis with the actual NLU hypothesis of the $u'$ from the pair $(u, u')$ in the annotated test set. For each given query, we retrieve the top 20 rewrites. These rewrites' NLU hypothesis are used to measure the system performance by standard information retrieval metrics: Precision@N (p@n) and Mean Reciprocal Rank(MRR). The p@n measures if at least one rewrite in the to $n$ candidates has a NLU hypothesis matched the $u'$'s, and MRR is the average reciprocal rank where the first correct rewrite is encountered.

We train the baseline system using the embedder and methods described in Figure \ref{fig2} with only the QR set. We experimented several contextual word embedders and seq2vec pooling methods on the development set, and chose the following setup that achieved the best performance: BERT-base model for contextual word embedder and a multi-head self-attention for the seq2vec Pooling. Specific for the self-attention, we use a \textit{max-pooled scaled dot-product attention} $\mathbf{w} = \operatorname{softmax}\left( \frac{\operatorname{max}\left(\textit{linear}(\mathbf{X})(\textit{linear}(\mathbf{X}))^{\operatorname{T}} \right)}{\sqrt{d}} \right)$ for each head, where $\mathbf{X}$ are embeddings input to the pooling layer. 

For pre-training, we adopt the same model architecture as the baseline, and follow the methods described in Section \ref{sec_approach_pre_train} using the pre-training set. We also pre-train a model by jointly training with NLU hypotheses. We refer to the first result with pre-training as "pretrain text" and the second result as "pretrain textNLU". After the pre-training is done. We further use the QR set to fine-tune the pre-trained model. In the Section \ref{sec_results}, we will discuss and compare their results.

We use a mini-batch size of 256 and set a maximum utterance length of 25. The utterance embedding vector size is 768. We also use dropout for the final linear layers with a drop ratio pf 0.3. We use the Adam optimizer and set the training epoch as 20 with early stopping. For both pre-training and fine-tuning, we use an initial learning rate of $2\times10^{-5}$. All experiments are done by PyTorch 1.1 on Nvidia V100 GPUs.



\vspace{-3mm}
\subsection{Experimental Results}
\label{sec_results}
\vspace{-1mm}
We summarize the experiments in Table \ref{table_qr_main_results} \footnote{Beside the results Table \ref{table_qr_main_results}, we also experimented Deep Structured Semantic Models (DSSM) \cite{huang2013learning} on the QR set, and obtained a p@1 of 0.16 and p@5 of 0.44, which is far behind the baseline established here.} to demo the effectiveness of pre-training. Comparing "baseline" with "pretrain text" and "pretrain textNLU", we see that pre-trained models already achieve very competitive results without any fine-tuning on the QR set. "pretrain textNLU" has a further 2\% gain in p@1 and 1\% gain in p@5 over "pretrain text". We then fine-tune the pre-trained models on 20\% of the QR set for just 2 epochs, and the result out-performs the baseline, which is fully trained on the entire QR set, by around 7\% or around 5\% with or without NLU hypotheses in pre-training.

\small{
\begin{table}[]
\captionsetup{skip=0pt}
\begin{tabular}{@{}llllll@{}}
\toprule
& \textbf{p@1}  & \textbf{p@5}  & \textbf{p@10} & \textbf{p@20} & \textbf{MRR}   \\ \midrule
\textbf{baseline}           & 0.293 & 0.477 & 0.569  & 0.669  & 16.05 \\
\textbf{pretrain text}      & 0.284 & 0.476 & 0.560  & 0.640  & 17.46 \\
\textbf{pretrain text\/NLU} & 0.290 & 0.481 & 0.563  & 0.638  & 17.19 \tabularnewline  \midrule
\multicolumn{6}{c}{after fine-tuning on 20\% QR set for 2 epochs} \tabularnewline \midrule
\shortstack{\textbf{pretrain text} \\ {}} & \shortstack{0.307 \\ {\small +4.8\%$^*$}} & \shortstack{0.505 \\ {\small +5.9\%}} & \shortstack{0.599 \\ {\small +5.3\%}} & \shortstack{0.689 \\ {\small +3.0\%}}& 15.13 \\
\shortstack{\textbf{pretrain text\/NLU} \\ {}} & \shortstack{0.315 \\ {\small +7.5\%}} & \shortstack{0.513 \\ {\small +7.5\%}} & \shortstack{0.608 \\ {\small +6.9\%}}  & \shortstack{0.680 \\ {\small +1.6\%} }  & 15.18
\tabularnewline \bottomrule
\end{tabular}
\vspace{2mm}
\caption{\small{Summary of QR experiment results. $^*$ relative performance gain in comparison to the "baseline".}}
\label{table_qr_main_results}
\vspace{-4mm}
\end{table}
}
\normalsize

\begin{figure}
    \centering
    \begin{subfigure}[t]{\textwidth}
        \raisebox{-\height}{\includegraphics[width=0.27\textwidth]{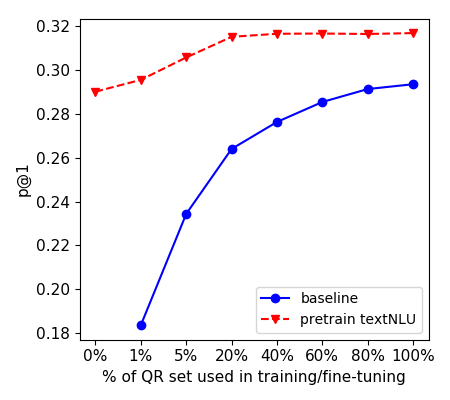}}\hspace{-0.7em}
        \raisebox{-\height}{\includegraphics[width=0.27\textwidth]{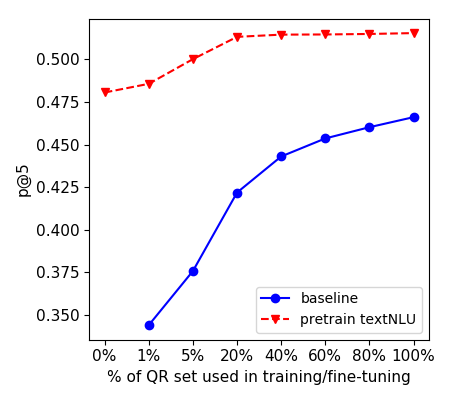}}
    \end{subfigure}
    \vspace{-2mm}
    \caption{\small{P@1 and p@5 performance with varying ratio of QR set used in training or fine-tuning.}}
    \label{fig_qr_train_ratio}
    \vspace{-5mm}
\end{figure}

A performance analysis by adjusting the ratio of the QR set used in fine-tuning is shown in Figure \ref{fig_qr_train_ratio}, where we compare the p@1 and p@5 between "baseline" and the "pretrain textNLU" after fine-tuning. We observe a definite advantage of the model with pre-training against the "baseline", especially when the training data size is smaller. Although the performance gap narrows when more QR data is included in training, the model with pre-training still gets a distinctive upper hand. Overall, the results also support our claim that pre-training provides rich prior information, so the model is almost saturated with 20\% of the QR set; in comparison, the baseline continues to gain with a larger training set.

\vspace{-3mm}
\section{Conclusion}
\vspace{-1mm}
This work proposes a neural retrieval based query rewriting approach to reduce user friction in a spoken language understanding system. The proposed neural QR model leverages a pre-trained word embedder like BERT. We further propose a pre-training procedure that makes use of a large amount of readily available historical user queries and their NLU hypotheses. The experiment results show the distinct advantage of pre-training in comparison to the QR model without pre-training - not just significantly reducing the requirement of high-quality QR training pairs, but also remarkably improving performance. While we focus on pre-training for QR task in this paper, we believe a similar strategy could potentially apply to other tasks in NLU, e.g. domain classification, NLU n-best ranking, etc.

\vfill\pagebreak

\bibliography{refs} 
\bibliographystyle{IEEEbib}

\end{document}